\documentclass[11pt]{article}

\usepackage{acl2015}
\usepackage{times}
\usepackage{url}
\usepackage{latexsym}
\usepackage{amssymb,amsmath,amsthm}
\usepackage{verbatim}
\usepackage[numbers,sort&compress]{natbib}
\usepackage{graphicx}
\usepackage{tikz}
\usetikzlibrary{arrows.meta}
\usepackage{capt-of}
\usepackage{xcolor}
\usepackage{microtype}
\usepackage{bm}
\usepackage{booktabs,tabularx,array}
\usepackage{enumitem}
\usepackage{xurl}
\usepackage{balance}
\usepackage{placeins}
\usepackage{hyperref}

\definecolor{linkblue}{RGB}{25,72,120}
\hypersetup{
  colorlinks=true,
  linkcolor=linkblue,
  citecolor=linkblue,
  urlcolor=linkblue,
  pdftitle={Semantics at an Angle: When Cosine Similarity Works Until It Doesn't},
  pdfauthor={Kisung You},
  pdfsubject={A technical review of cosine similarity for learned representations},
  pdfkeywords={cosine similarity, embeddings, vector norms, anisotropy, hubness, retrieval}
}

\setlist[itemize]{leftmargin=*,topsep=3pt,itemsep=2pt}
\setlist[enumerate]{leftmargin=*,topsep=3pt,itemsep=2pt}
\newcommand{\R}{\mathbb{R}}
\newcommand{\Sph}{\mathbb{S}}
\newcommand{\cossim}{\operatorname{cos}}
\newcommand{\norm}[1]{\left\lVert #1 \right\rVert}

\title{Semantics at an Angle: When Cosine Similarity Works Until It Doesn't}
\date{Revised July 2026}

\author{
Kisung You\\
Department of Mathematics\\
Baruch College, City University of New York\\
\href{mailto:kisung.you@baruch.cuny.edu}{kisung.you@baruch.cuny.edu}\\[0.4ex]
\small Revised July 2026
}

\begin{document}
\maketitle
\pagestyle{plain}
\raggedbottom

\begin{abstract}
Cosine similarity is a standard comparison rule for learned representations in
information retrieval, natural language processing, computer vision, and
multimodal learning. Its popularity is well founded: it removes positive radial
scale, is computationally convenient, and often matches objectives that train
normalized embeddings. These same properties also delimit what cosine can
express. Normalization discards radial variation; anisotropic representations
can compress angular contrast; high-dimensional neighborhoods can develop
hubs; and a symmetric, uncalibrated score may be mismatched to the relation of
interest. This article offers a selective review organized around a simple
principle: the usefulness of cosine similarity depends jointly on the learned
representation, any normalization or post-processing, the scoring rule, and the
downstream decision. We derive the main geometric identities, distinguish
failure mechanisms that are often conflated, review representative evidence
about embedding norms, and describe objective-matched, geometry-aware,
hubness-aware, norm-aware, and learned alternatives. The central conclusion is
conditional rather than adversarial: cosine's positive-scale invariance is
justified when radial variation is nuisance or fixed by the representation
contract, but its angular geometry and downstream decision must still be
validated.
\end{abstract}


\section{Introduction}

Learned embeddings turn objects into vectors so that a downstream system can
compare, retrieve, cluster, or classify them. Once an encoder
$E:\mathcal{X}\rightarrow\R^d$ has been trained, it is tempting to regard the
choice of comparison rule as a minor implementation detail. In practice, that
choice imposes an additional modeling assumption. For nonzero vectors
$\bm{x}=E(x)$ and $\bm{y}=E(y)$, cosine similarity retains only their
directions. It therefore treats $\bm{x}$ and $a\bm{x}$ as equivalent for every
$a>0$, regardless of whether their different lengths arose from nuisance
scale, confidence, frequency, input complexity, or task-relevant signal.

This observation does not imply that cosine similarity is generally defective.
In sparse document retrieval, scale invariance attenuates document-length
effects. In contrastive learning, embeddings are often normalized within the
training objective, making a spherical comparison part of the model itself
\citep{chen2020simclr,wang2020alignment,radford2021clip}. For any unit-normalized
representation, cosine similarity, inner product, and Euclidean distance induce
the same nearest-neighbor ranking. Under those conditions, cosine is not merely
a convenient default; it is an objective-matched score.

\begin{figure*}[t]
  \centering
  \includegraphics[width=\textwidth]{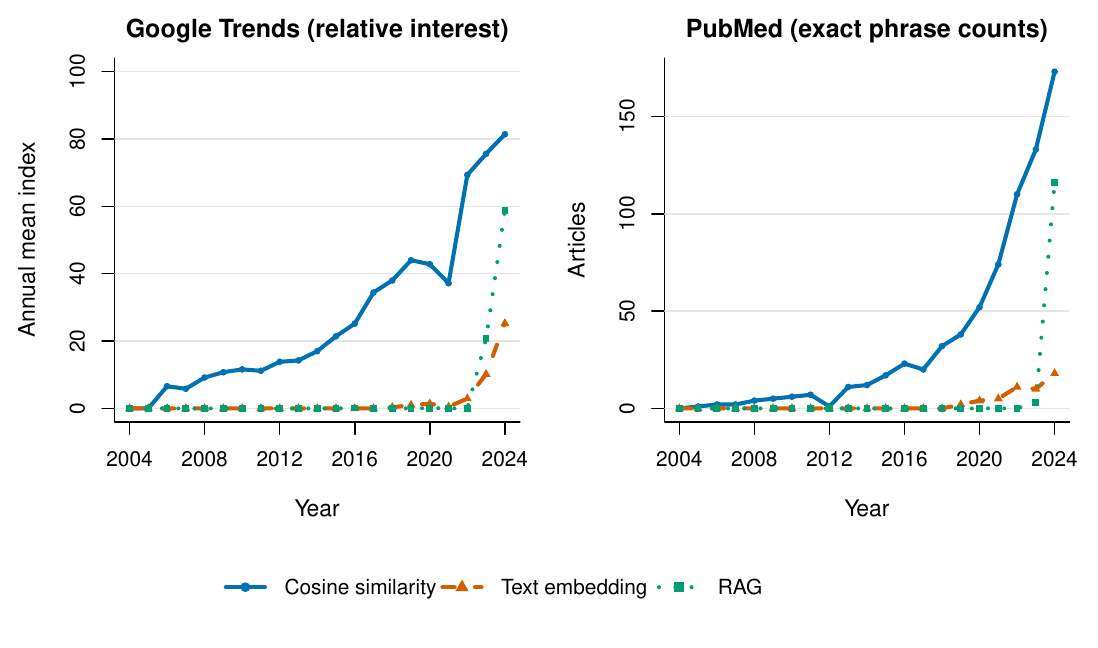}
  \caption{Descriptive attention to three embedding-related search terms,
  2004--2024. Left: annual means from one joint worldwide Google Trends
  comparison of quoted phrases (relative 0--100 interest). Right: yearly PubMed counts for
  the exact phrases in Title/Abstract under the Publication Date filter,
  refreshed July 23, 2026. Trends values are sampled and normalized, so zeros
  may reflect low volume. The series measure attention to query strings, not
  method quality; acquisition details accompany the source.}
  \label{fig:trends}
\end{figure*}

The surrounding embedding vocabulary has also become substantially more
visible. Figure~\ref{fig:trends} provides descriptive context by comparing
search attention and publication counts for cosine similarity, text embedding,
and retrieval-augmented generation. These curves measure attention to specific
query strings, not technical merit or comparative accuracy.

Problems arise when this model contract is absent or misunderstood. The norm of
a learned vector can contain information in some models
\citep{oyama2023norm,draganov2025norms}; the directions of contextual
representations can be strongly anisotropic \citep{ethayarajh2019contextual};
and nearest-neighbor graphs can exhibit hubness even when all points have unit
norm \citep{radovanovic2010hubness,bogolin2022qbnorm}. These are distinct
phenomena. Recovering norms cannot by itself repair anisotropy or hubness, and
making a representation isotropic does not make a symmetric score suitable for
an asymmetric relation such as entailment.

The aim of this article is to make those distinctions precise while retaining a
compact, tutorial style. It is a selective review rather than an exhaustive
survey or systematic evidence synthesis: sources are selected to illustrate
mechanisms and remedies, not to estimate an average effect across models. We
emphasize foundational work and representative evidence available through
2025, with examples from text, vision, multimodal learning, and dense retrieval.
Our contributions are fourfold:

\begin{itemize}
  \item We state the geometric properties of cosine similarity carefully,
  including its invariances, its relationship to metric distances on the
  sphere, and the role of vector norm in optimization.
  \item We explain historically why cosine became a default while separating
  the training objective from the downstream evaluation score.
  \item We organize limitations into four operational diagnoses: radial
  information loss, global geometric distortion (including anisotropy and
  concentration), local-density or hubness effects, and task--score mismatch.
  We treat representation--score non-identifiability as a cross-cutting warning
  about interpreting learned coordinates.
  \item We give a practical decision procedure for choosing among cosine,
  inner product, post-processing, hubness correction, norm-aware scoring, and
  learned comparison functions.
\end{itemize}

The resulting thesis is deliberately conditional:
\emph{cosine similarity quotients out positive radial scale. That invariance is
justified when radial scale is nuisance or fixed by the representation
contract; it is not by itself sufficient to make cosine the right score for the
downstream decision.}

\section{Four Objects That Should Not Be Conflated}
\label{sec:four_objects}

Claims that cosine works or fails are incomplete unless they specify what
produced the vectors and how the scores are used. A useful analysis separates
four objects.

\begin{table*}[t!]
\centering
\caption{Four levels in an embedding comparison pipeline. A failure observed at
one level should not automatically be attributed to another.}
\label{tab:four_levels}
\small
\begin{tabularx}{\textwidth}{
  >{\raggedright\arraybackslash}p{0.16\textwidth}
  >{\raggedright\arraybackslash}p{0.25\textwidth}
  >{\raggedright\arraybackslash}X}
\toprule
\textbf{Level} & \textbf{Mathematical object} & \textbf{Question} \\
\midrule
Representation &
$E(x)\in\R^d$ &
What information and invariances did the training objective encode? \\
Transformation &
$T(E(x))$, such as normalization, centering, or whitening &
Which variation is removed, rescaled, or estimated from a reference corpus? \\
Pairwise score &
$s(T(E(x)),T(E(y)))$ &
Does the comparison use direction, norm, local density, or learned interactions? \\
Decision &
ranking, threshold, clustering, or calibrated probability &
What task loss is optimized, and how stable is the rule across domains? \\
\bottomrule
\end{tabularx}
\end{table*}

Table~\ref{tab:four_levels} is the organizing frame used throughout this
review. For example, anisotropy is primarily a property of a representation
distribution, although post-processing can change it. Hubness is a property of
the induced neighborhood graph, not an attribute stored in the norm of one
vector. Calibration is a property of scores relative to outcomes and a target
population. Cosine normalization, in contrast, is a deterministic
transformation followed by an inner product. Keeping these levels separate
prevents a single angle-versus-norm story from being asked to explain every
failure.

\subsection{A cross-cutting identifiability warning}

Even without normalization, the geometry of learned factors need not be
uniquely identified by the data-fit term. Consider an unconstrained two-factor
model whose predictions depend only on $\bm{X}\bm{Y}^{\mathsf T}$. For any
invertible matrix $\bm{A}$,
\begin{equation}
  (\bm{X}\bm{A})(\bm{Y}\bm{A}^{-\mathsf T})^{\mathsf T}
  =\bm{X}\bm{Y}^{\mathsf T}.
  \label{eq:gauge}
\end{equation}
Equation~\eqref{eq:gauge} preserves this product and hence the model's
predictions, but cosines within the transformed factor spaces generally change
unless $\bm{A}$ is a scaled orthogonal transformation. The full invariance need
not survive tied parameterizations, normalization, explicit geometric
constraints, or regularization; such choices can restrict or select among
equivalent factorizations. \citet{steck2024cosine} analyze how regularization
controls this issue in linear embedding models. The applicable lesson is
conditional: semantic interpretation requires checking the invariances of the
actual training objective, not only the coordinates it outputs.

\paragraph{Terminology.}
We use \emph{representation contract} for the combination of exposed
representation, normalization, and pairwise interaction that a downstream user
is expected to apply. A \emph{unit-normalized interface} is one such contract.
Throughout, \emph{spherical} means that the compared embeddings have fixed norm
(unit norm unless stated otherwise). It does not imply \emph{isotropy}, which
is a distributional property: ideally, a random direction $\bm{U}$ is isotropic
if $\bm{Q}\bm{U}\overset{d}{=}\bm{U}$ for every orthogonal matrix $\bm{Q}$.
Empirical anisotropy diagnostics measure departures from this ideal.

\FloatBarrier

\section{Mathematical Anatomy}
\label{sec:mathematics}

\subsection{Direction, radius, and invariance}

Let $\bm{x},\bm{y}\in\R^d\setminus\{\bm{0}\}$. Write each vector in polar form,
\begin{equation}
  \begin{aligned}
    \bm{x} &= r_x\bm{u}_x, \\
    r_x &= \norm{\bm{x}}>0, \\
    \bm{u}_x &= \frac{\bm{x}}{\norm{\bm{x}}}\in\Sph^{d-1}.
  \end{aligned}
  \label{eq:polar}
\end{equation}
and analogously for $\bm{y}$. Cosine similarity is
\begin{equation}
  \cossim(\bm{x},\bm{y})
  =\frac{\bm{x}^{\mathsf T}\bm{y}}
         {\norm{\bm{x}}\norm{\bm{y}}}
  =\bm{u}_x^{\mathsf T}\bm{u}_y
  =\cos\theta,
  \label{eq:cosine}
\end{equation}
where $\theta\in[0,\pi]$ is the angle between the vectors. It is undefined when
either vector is zero and takes values in $[-1,1]$.

For $a,b>0$,
\begin{equation}
  \cossim(a\bm{x},b\bm{y})=\cossim(\bm{x},\bm{y}).
  \label{eq:scale_invariance}
\end{equation}
Thus cosine is invariant to \emph{positive} rescaling. Negative rescaling flips
direction and, consequently, the sign of the score. Geometrically,
normalization maps each positive ray in $\R^d\setminus\{\bm{0}\}$ to one point
on $\Sph^{d-1}$. Every value carried only by $r_x$ is removed from the
downstream score.

\subsection{Cosine is a similarity, not a metric}

Cosine similarity is not a distance metric. The commonly used cosine
dissimilarity
\begin{equation}
  d_{\mathrm{cos}}(\bm{x},\bm{y})
  =1-\cossim(\bm{x},\bm{y})
\end{equation}
is nonnegative and symmetric. On $\R^d\setminus\{\bm{0}\}$, however, it assigns
zero to distinct vectors on the same positive ray, and it can violate the
triangle inequality. Restricting the domain to the unit sphere repairs identity
of indiscernibles but not the triangle inequality. Two closely related
quantities do define metrics on the unit sphere:
\begin{align}
  d_{\mathrm{ang}}(\bm{x},\bm{y})
    &=\arccos\!\left(\cossim(\bm{x},\bm{y})\right),\\
  d_{\mathrm{chord}}(\bm{x},\bm{y})
    &=\norm{\bm{u}_x-\bm{u}_y}
      =\sqrt{2-2\cossim(\bm{x},\bm{y})}.
  \label{eq:spherical_distances}
\end{align}
The first is the geodesic angular distance and the second is the Euclidean
chord distance. Both are strictly decreasing transformations of cosine
similarity and therefore reverse, but otherwise preserve, pairwise rankings.
The von Mises--Fisher family provides a probabilistic model whose log density
depends on an inner product with a mean direction
\citep{fisher1953dispersion}, but this does not make raw cosine scores
calibrated probabilities.

\subsection{Equivalence after normalization}

The squared Euclidean distance decomposes as
\begin{align}
  \norm{\bm{x}-\bm{y}}^2
  &=(r_x-r_y)^2
    +2r_xr_y\{1-\cossim(\bm{x},\bm{y})\}.
  \label{eq:l2_decomposition}
\end{align}
When $r_x=r_y=1$,
\begin{equation}
  \begin{aligned}
    \bm{x}^{\mathsf T}\bm{y}
      &=\cossim(\bm{x},\bm{y}), \\
    \norm{\bm{x}-\bm{y}}^2
      &=2-2\cossim(\bm{x},\bm{y}).
  \end{aligned}
  \label{eq:normalized_equivalence}
\end{equation}
Consequently, cosine similarity, dot product, and Euclidean distance produce
identical rankings over a collection of unit vectors. They may still yield
different numerical thresholds or calibration maps, so ranking equivalence
should not be confused with score equivalence.

Figure~\ref{fig:cosine_geometry} summarizes these relationships. Cosine fixes
only the angle $\theta$; ordinary Euclidean distance additionally depends on
$r_x$ and $r_y$, while normalization replaces the original endpoints by
$\bm{u}_x$ and $\bm{u}_y$ and makes their Euclidean separation a chord of the
unit sphere.

\par\medskip
\noindent\begin{minipage}{\columnwidth}
  \centering
  \begin{tikzpicture}[
  x=2cm,
  y=2cm,
  >=Latex,
  line cap=round,
  line join=round,
  every node/.style={font=\scriptsize}
]
  \coordinate (O)  at (0,0);
  \coordinate (Ux) at (0:1);
  \coordinate (Uy) at (52:1);
  \coordinate (X)  at (0:1.72);
  \coordinate (Y)  at (52:1.48);

  \draw[gray!55] (O) circle[radius=1];
  \node[text=gray!70,anchor=east] at (215:1.03) {$\Sph^{d-1}$};
  \draw[->,semithick] (O) -- (X);
  \draw[->,semithick] (O) -- (Y);

  \draw[very thick] (Ux) -- (Uy)
    node[midway,anchor=east,xshift=-1pt] {$d_{\mathrm{chord}}$};
  \draw[thick,densely dashed] (X) -- (Y)
    node[midway,anchor=west,xshift=1pt] {$d_{\mathrm E}$};

  \draw (0.34,0) arc[start angle=0,end angle=52,radius=0.34];
  \node at (26:0.46) {$\theta$};

  \fill (O)  circle[radius=1pt];
  \fill (Ux) circle[radius=1.2pt];
  \fill (Uy) circle[radius=1.2pt];
  \node[anchor=north east] at (O)  {$\bm{0}$};
  \node[anchor=north]      at (Ux) {$\bm{u}_x$};
  \node[anchor=south east] at (Uy) {$\bm{u}_y$};
  \node[anchor=west]       at (X)  {$\bm{x}$};
  \node[anchor=south west] at (Y)  {$\bm{y}$};

  \node[anchor=west] at (-0.97,-1.20)
    {$\cossim(\bm{x},\bm{y})
      =\bm{u}_x^{\mathsf T}\bm{u}_y=\cos\theta$};
  \end{tikzpicture}
\captionof{figure}{Cosine, chordal distance, and Euclidean distance in the plane spanned
by $\bm{x}=r_x\bm{u}_x$ and $\bm{y}=r_y\bm{u}_y$. On the unit sphere,
$d_{\mathrm{chord}}=\norm{\bm{u}_x-\bm{u}_y}
=\sqrt{2-2\cos\theta}$. For the original vectors,
$d_{\mathrm E}^2=\norm{\bm{x}-\bm{y}}^2
=(r_x-r_y)^2+2r_xr_y(1-\cos\theta)$. Thus Euclidean distance also depends on
the radii; for unit vectors, it equals the chordal distance.}
\label{fig:cosine_geometry}
\end{minipage}
\par\medskip

Table~\ref{tab:measures} makes the domain restrictions and retained information
of the main comparison rules explicit.

\begin{table*}[t!]
\centering
\caption{Common comparison rules. Metric refers to the mathematical distance
axioms on the stated domain.}
\label{tab:measures}
\small
\begin{tabularx}{\textwidth}{
  >{\raggedright\arraybackslash}p{0.17\textwidth}
  >{\raggedright\arraybackslash}p{0.25\textwidth}
  >{\centering\arraybackslash}p{0.11\textwidth}
  >{\centering\arraybackslash}p{0.12\textwidth}
  >{\raggedright\arraybackslash}X}
\toprule
\textbf{Rule} & \textbf{Definition} & \textbf{Uses norm?} &
\textbf{Metric?} & \textbf{Key property} \\
\midrule
Cosine similarity &
$\bm{x}^{\mathsf T}\bm{y}/(r_xr_y)$ &
No &
No &
Invariant to positive rescaling. \\
Inner product &
$\bm{x}^{\mathsf T}\bm{y}=r_xr_y\cos\theta$ &
Yes &
No &
Combines angular alignment with both radii. \\
Euclidean distance &
$\norm{\bm{x}-\bm{y}}$ &
Yes &
Yes &
Sensitive to radial and angular differences. \\
Angular distance &
$\arccos(\cossim(\bm{x},\bm{y}))$ &
No &
Yes, on $\Sph^{d-1}$ &
Geodesic distance between directions. \\
Chord distance &
$\norm{\bm{u}_x-\bm{u}_y}$ &
No &
Yes, on $\Sph^{d-1}$ &
Euclidean metric with the same ranking as angular distance. \\
\bottomrule
\end{tabularx}
\end{table*}

\subsection{Norms can affect learning even when the score ignores them}

Cosine similarity is radially invariant as a function value, but its gradient
with respect to an unnormalized embedding depends on that embedding's norm:
\begin{equation}
  \nabla_{\bm{x}}\cossim(\bm{x},\bm{y})
  =\frac{1}{\norm{\bm{x}}}
   \left(\bm{I}-\bm{u}_x\bm{u}_x^{\mathsf T}\right)\bm{u}_y.
  \label{eq:cos_gradient}
\end{equation}
The gradient is tangent to the sphere because it is orthogonal to $\bm{x}$, and
its magnitude is
$\norm{\nabla_{\bm{x}}\cossim(\bm{x},\bm{y})}
=\sin\theta/\norm{\bm{x}}$.
Thus it has inverse-radius scaling at a fixed nondegenerate angle but vanishes
at exact alignment or anti-alignment. Pre-normalization radii can therefore
affect optimization speed even though they do not affect the forward cosine
score. Recent analysis of self-supervised learning makes this interaction
explicit and also reports model-dependent relationships between
pre-normalization norm and confidence \citep{draganov2025norms}. This
distinction between \emph{the vector scored downstream} and \emph{the vector
that generated the training gradient} is essential.

\section{How Cosine Became a Default}
\label{sec:history}

\subsection{Sparse information retrieval}

The SMART system was already being used to evaluate automatic indexing and
retrieval procedures in the 1960s \citep{salton1965smart}. An influential
later account of the vector-space model represented documents and queries by
weighted term vectors and described angular proximity as a comparison rule
\citep{salton1975vector}. In this setting, raw magnitude is strongly
influenced by document length and total term count. Angular normalization
offered a direct way to compare term distributions without allowing long
documents to dominate solely because they contained more tokens. This was an
appropriate invariance for many sparse retrieval problems, although it did not
remove the need for careful term weighting.

Cosine was not the only retrieval principle, and later probabilistic systems
should not be described as naive length-sensitive competitors. BM25, developed
after the early vector-space work, explicitly combines term saturation,
inverse-document-frequency weighting, and document-length normalization
\citep{robertson2009probabilistic}. The historical lesson is therefore not that
cosine solved a defect ignored by every alternative. Rather, several retrieval
traditions encoded different assumptions about frequency, length, and
relevance.

\subsection{Static and sentence embeddings}

Dense word embeddings strengthened the association between semantic relatedness
and local vector geometry. Skip-gram with negative sampling is trained with
word--context dot products \citep{mikolov2013word2vec,mikolov2013distributed},
while GloVe fits a weighted model of log co-occurrence counts
\citep{pennington2014glove}. Cosine
became a common \emph{evaluation} rule because it suppresses frequency- and
scale-related variation and often gives useful neighborhoods. It is important,
however, not to equate the downstream cosine rule with the training objective:
the latter can depend on both direction and magnitude.

The familiar word-analogy calculation
\[
  \bm{v}_{\texttt{king}}-\bm{v}_{\texttt{man}}
  +\bm{v}_{\texttt{woman}}
\]
uses cosine or another nearest-neighbor rule to select a candidate, but the
relational structure is learned by the embedding model; it is not created by
cosine similarity itself. Sentence-BERT later trained siamese encoders to
produce sentence vectors intended for efficient cosine-based inference
\citep{reimers2019sentencebert}. Its regression variant optimizes cosine
directly, whereas its classification and triplet variants use different
objectives; train--score alignment is therefore variant-specific. SimCSE made
the cosine contract explicit for sentence embeddings by optimizing a
contrastive objective over normalized representations
\citep{gao2021simcse}.

\subsection{Contrastive, non-contrastive, and multimodal learning}

Modern representation learning gives cosine a more structural role. SimCLR
normalizes projection-head outputs and uses a temperature-scaled contrastive
objective \citep{chen2020simclr}. MoCo uses a related contrastive formulation
with normalized representations \citep{he2020moco}. CLIP and ALIGN learn joint
image--text spaces with normalized cross-modal scores and a learned or tuned
temperature \citep{radford2021clip,jia2021align}. On the sphere, contrastive
learning can be interpreted through alignment of positive pairs and uniformity
of the representation distribution \citep{wang2020alignment}.

Not every self-supervised method relies on negative examples. BYOL compares
normalized predictions and targets while avoiding explicit negative pairs
\citep{grill2020byol}. This matters conceptually: normalization and cosine-based
alignment do not uniquely imply a contrastive loss. The training mechanism,
stop-gradient operation, predictor architecture, and score must be discussed
separately.

Nor is dense retrieval universally cosine-based. Dense Passage Retrieval
trains dual encoders with an inner-product score \citep{karpukhin2020dpr}.
Across modern systems, cosine is best understood as one common model contract
among several, not as the inevitable definition of semantic proximity.

\subsection{Why it often works}

The historical record suggests four recurring reasons for cosine's
effectiveness:

\begin{enumerate}
  \item \textbf{Useful invariance.} Positive radial scale is frequently a
  nuisance generated by document length, activation scale, or encoder
  parameterization.
  \item \textbf{Objective alignment.} When training normalizes embeddings,
  angular comparison is part of the learned geometry.
  \item \textbf{Computational convenience.} Unit vectors reduce cosine search
  to inner-product search and support efficient nearest-neighbor indexing.
  \item \textbf{Interpretability.} Direction and angle provide an intuitive,
  bounded description of alignment.
\end{enumerate}

None of these reasons establishes universal optimality. Each is a hypothesis
about the representation and task.

\section{What Can an Embedding Norm Mean?}
\label{sec:norm}

Equation~\eqref{eq:polar} separates radius from direction, but it does not assign
semantics to either component. Any interpretation of $r_x$ must be established
from the model or from controlled evidence.

\subsection{Representative evidence}

Table~\ref{tab:norm_evidence} collects the following studies as selected
evidence, not as a systematic estimate of how often norms are useful.

\begin{table*}[t!]
\centering
\caption{Selected evidence about embedding norms and cosine behavior. Findings
are conditional on the listed representation and evaluation setting.}
\label{tab:norm_evidence}
\small
\begin{tabularx}{\textwidth}{
  >{\raggedright\arraybackslash}p{0.18\textwidth}
  >{\raggedright\arraybackslash}p{0.22\textwidth}
  >{\raggedright\arraybackslash}p{0.34\textwidth}
  >{\raggedright\arraybackslash}X}
\toprule
\textbf{Study} & \textbf{Representation} & \textbf{Reported finding} &
\textbf{Qualification} \\
\midrule
\citet{oyama2023norm} &
Static skip-gram word embeddings &
Squared norm is connected to an information-gain quantity and is not explained
only by word frequency in the reported experiments. &
The derivation and evidence are objective- and model-specific. \\
\citet{amigo2025correspondence} &
Contextual and composed text embeddings &
Norm--information correspondence changes with the subword representation and
text-composition rule. &
The correspondence is not automatic for pooled outputs. \\
\citet{zhou2022frequency} &
Contextual BERT word representations &
Cosine underestimates human-rated similarities involving high-frequency words
in the studied settings. &
The result does not establish the same bias for every encoder or layer. \\
\citet{draganov2025norms} &
Self-supervised vision representations &
Pre-normalization norm affects cosine-gradient dynamics and relates to
confidence or sample unexpectedness in the studied models. &
The evidence concerns pre-normalization, model-dependent quantities. \\
\bottomrule
\end{tabularx}
\end{table*}

For static word embeddings trained with skip-gram negative sampling,
\citet{oyama2023norm} connect squared norm to an information-gain quantity
derived from the word's co-occurrence distribution. They also provide
experiments designed to separate that relationship from simple word-frequency
correlation. In contextual BERT representations, \citet{zhou2022frequency}
show that cosine similarity systematically underestimates similarities
involving frequent words relative to human judgments, even after controlling
for several confounders. Their geometric explanation is informative but should
not be generalized to every contextual encoder without verification.

In self-supervised vision, \citet{draganov2025norms} show that
pre-normalization norms influence cosine-gradient dynamics and, in their
studied settings, relate to confidence or sample unexpectedness. These results
are strong evidence that a radius discarded by the forward score can still
matter during learning.

\citet{amigo2025correspondence} extend the norm--information question to
contextual and composed text embeddings. Their results show that the
correspondence depends strongly on the subword representation and composition
rule rather than holding automatically for pooled outputs.

\subsection{Why the interpretation is conditional}

The same norm can also encode unwanted effects: input length, token frequency,
modality-specific scale, pooling choice, layer normalization, or domain shift.
Furthermore, a deployed system may expose only unit-normalized outputs, in
which case every downstream norm is identically one and the informative
pre-normalization radius is unavailable. Evidence about a hidden representation
does not automatically justify a norm-aware score on the released
representation.

A careful analysis should therefore ask:

\begin{itemize}
  \item Which layer and which side of the normalization operation supplies the
  vector?
  \item Does norm predict a task outcome after controlling for length,
  frequency, class, and domain?
  \item Does a norm-aware score improve held-out ranking or calibration, rather
  than merely correlate in-sample?
  \item Is the relationship stable across model checkpoints and deployment
  populations?
\end{itemize}

The defensible claim is not that norm is always signal. It is that
\emph{norm is a model-dependent variable whose removal should be tested when
the representation contract does not already declare it irrelevant.}

\section{A Taxonomy of Failure Mechanisms}
\label{sec:failures}

The four operational diagnoses require different evidence and different first
responses. Table~\ref{tab:failure_taxonomy} provides the compact map used in
this section.

\begin{table*}[t!]
\centering
\caption{Failure mechanisms, diagnostics, and targeted responses.}
\label{tab:failure_taxonomy}
\small
\begin{tabularx}{\textwidth}{
  >{\raggedright\arraybackslash}p{0.18\textwidth}
  >{\raggedright\arraybackslash}p{0.27\textwidth}
  >{\raggedright\arraybackslash}p{0.25\textwidth}
  >{\raggedright\arraybackslash}X}
\toprule
\textbf{Mechanism} & \textbf{Diagnostic} & \textbf{Candidate response} &
\textbf{Main caveat} \\
\midrule
Radial information loss &
Norm--outcome association after confounder adjustment; cosine versus dot-product ablation &
Inner product, explicit norm features, learned angular--radial score &
Norm may be nuisance or unavailable after normalization. \\
Global geometric distortion &
Mean direction, covariance spectrum, pairwise-score spread, neighbor-score margins, layerwise comparison &
Centering, component removal, regularized whitening, dimensionality reduction, learned transformation &
The remedy is distribution-dependent and may remove useful signal. \\
Local density and hubness &
$N_k$ distribution, skewness, neighbor overlap, cross-modal imbalance &
CSLS, querybank normalization, reciprocal or local-density ranking &
Requires a representative reference bank and extra computation. \\
Task--score mismatch &
Compare task metrics, symmetry requirements, reliability diagrams, domain shifts &
Objective-matched score, bilinear model, calibration, reranker or cross-encoder &
More flexible scores require labels, computation, or both. \\
\bottomrule
\end{tabularx}
\end{table*}

\subsection{Radial information loss}

Suppose $\bm{x}_1=r_1\bm{u}$ and $\bm{x}_2=r_2\bm{u}$ with $r_1\neq r_2$.
For every nonzero $\bm{y}$,
\begin{equation}
  \cossim(\bm{x}_1,\bm{y})
  =\cossim(\bm{x}_2,\bm{y}).
  \label{eq:radial_loss}
\end{equation}
No downstream procedure that receives only cosine scores can recover the
difference between $r_1$ and $r_2$. This is a true limitation precisely when
the radius predicts relevance, confidence, informativeness, or another target
after nuisance variables are controlled. If radius is arbitrary activation
scale, Equation~\eqref{eq:radial_loss} is a benefit instead.

\subsection{Global geometric distortion}

Global anisotropy is a departure of the directional distribution from the
rotationally invariant ideal. For unit vectors
$\{\bm{u}_i\}_{i=1}^n$ with mean $\bar{\bm{u}}$, the average off-diagonal
cosine is
\begin{equation}
  A_{\mathrm{cos}}
  =\frac{1}{n(n-1)}\sum_{i\neq j}\bm{u}_i^{\mathsf T}\bm{u}_j
  =\frac{n\norm{\bar{\bm{u}}}^2-1}{n-1}.
  \label{eq:anisotropy}
\end{equation}
A large positive value therefore diagnoses a common mean direction. It is not
a complete definition of isotropy: a zero-mean distribution can still have a
highly nonuniform covariance spectrum. Contextual word representations from
several pretrained models have been found to occupy narrow cones
\citep{ethayarajh2019contextual}, and anisotropic sentence spaces can degrade
raw similarity comparisons \citep{li2020bertflow}.

Anisotropy is not an inherent property of every Transformer.
\citet{machina2024anisotropy} identify large Pythia models with near-isotropic
representations and connect the result to architecture and final layer
normalization. Claims about anisotropy should therefore specify the model,
layer, pooling rule, corpus, and diagnostic.

\subsubsection{Distance concentration}

In generic high-dimensional settings, distance-based proximity can lose
relative contrast \citep{aggarwal2001distance}. For cosine on the unit sphere,
an analogous baseline follows directly from random directions. If $\bm{U}$ and
$\bm{V}$ are independent and uniform on $\Sph^{d-1}$, then
\begin{equation}
  \mathbb{E}[\bm{U}^{\mathsf T}\bm{V}]=0,\qquad
  \operatorname{Var}(\bm{U}^{\mathsf T}\bm{V})=\frac{1}{d}.
  \label{eq:spherical_concentration}
\end{equation}
Thus pairwise cosines concentrate around zero as dimension grows, while the
equivalent chord distances concentrate around $\sqrt{2}$.

Random-pair concentration alone does not show that semantic retrieval has
degraded: relevant pairs need not resemble independent uniform directions, and
corpus size and signal structure affect the nearest-neighbor tail. Diagnostics
should therefore include score margins or relative contrast in the actual
embedding corpus, not dimension alone.

\subsection{Local density and hubness}

Hubness is related to concentration but distinct. Given a query set $Q$, gallery
$G$, and
top-$k$ neighborhood $\mathcal{N}_k(q)$, define the $k$-occurrence of gallery
item $g$ by
\begin{equation}
  N_k(g)=\sum_{q\in Q}\mathbf{1}\{g\in\mathcal{N}_k(q)\}.
  \label{eq:hubness}
\end{equation}
Hubs have unusually large $N_k(g)$, producing a right-skewed occurrence
distribution \citep{radovanovic2010hubness}. They can arise with Euclidean or
angular neighborhoods and can be amplified when query and gallery
distributions differ, as in cross-modal retrieval \citep{bogolin2022qbnorm}.
Hubness is not simply information contained in norm, so restoring radii is not
a general remedy.

\subsection{Task and calibration mismatch}

Cosine is symmetric:
\[
  \cossim(\bm{x},\bm{y})=\cossim(\bm{y},\bm{x}).
\]
It therefore cannot by itself represent an intrinsically asymmetric relation
such as textual entailment, query-to-document utility, or directed
substitutability. A bi-encoder can still encode roles through different query
and document encoders, but the suitability of a symmetric interaction must be
established by training and evaluation.

Cosine is also bounded but uncalibrated. A score of $0.8$ is not a probability,
and a threshold learned for one model, layer, language, or corpus need not
transfer to another. Ranking quality, classification calibration, and semantic
interpretation are different evaluation targets.

\section{Targeted Remedies}
\label{sec:remedies}

\subsection{Start with the model contract}

The first alternative to blind cosine use is not a more elaborate metric; it is
to identify the score used during training.

\begin{itemize}
  \item If the model emits unit vectors and was trained with normalized
  similarity, cosine, dot product, and Euclidean distance have identical
  rankings by Equation~\eqref{eq:normalized_equivalence}. Choose among them for
  implementation convenience, then calibrate separately if thresholds matter.
  \item If the model was trained with an unnormalized inner product, compare
  inner product and cosine on held-out task metrics before discarding the
  learned radii.
  \item If the embeddings were not trained for pairwise comparison, treat every
  off-the-shelf similarity rule as a baseline rather than a semantic ground
  truth.
\end{itemize}

\subsection{Centering, component removal, and whitening}

When a common direction dominates, mean-centering replaces
$\bm{x}$ by $\bm{x}-\bm{\mu}$, where $\bm{\mu}$ is estimated on a reference
corpus. All-but-the-top additionally removes a small number of leading
principal directions \citep{mu2018allbuttop}. If the centered covariance is
$\bm{\Sigma}=\bm{U}\bm{\Lambda}\bm{U}^{\mathsf T}$, a regularized whitening
map has the form
\begin{equation}
  T_{\mathrm{white}}(\bm{x})
  =(\bm{\Lambda}+\varepsilon\bm{I})^{-1/2}
    \bm{U}^{\mathsf T}(\bm{x}-\bm{\mu}),
  \label{eq:whitening}
\end{equation}
possibly followed by dimensionality reduction and unit normalization.
Whitening-based methods have improved unsupervised sentence embeddings in
specific BERT settings \citep{huang2021whiteningbert}; normalizing-flow
transformations provide a nonlinear alternative \citep{li2020bertflow}.

These are estimated transformations, not universal repairs. The reference
corpus must represent deployment conditions; covariance estimation may require
regularization; and flattening high-variance directions can erase task signal.
Every post-processing step should be fitted without test leakage and evaluated
against an unprocessed baseline.

\subsection{Local-density and querybank corrections}

Cross-domain similarity local scaling (CSLS) adjusts a raw score using local
neighborhood averages \citep{conneau2018csls}. For cosine or another base score
$s(q,g)$, let
\begin{align}
  r_G(q)&=\frac{1}{k}\sum_{g'\in\mathcal{N}_G(q)}s(q,g'),\\
  r_Q(g)&=\frac{1}{k}\sum_{q'\in\mathcal{N}_Q(g)}s(q',g).
\end{align}
Then
\begin{equation}
  s_{\mathrm{CSLS}}(q,g)
  =2s(q,g)-r_G(q)-r_Q(g).
  \label{eq:csls}
\end{equation}
The gallery-dependent term penalizes items that are broadly similar to many
queries.

Querybank normalization generalizes this idea for cross-modal retrieval
\citep{bogolin2022qbnorm}. Given a bank of representative queries
$B=\{b_i\}_{i=1}^m$ and gallery items $\{g_j\}$, precompute the probe matrix
$P_{ji}=s(b_i,g_j)$. One option, inverted softmax, uses
\begin{equation}
  s_{\mathrm{IS}}(q,g_j)
  =
  \frac{\exp\{\beta s(q,g_j)\}}
       {\sum_{i=1}^{m}\exp\{\beta P_{ji}\}},
  \label{eq:inverted_softmax}
\end{equation}
where $\beta>0$ is an inverse temperature. The denominator depends on the
gallery item $g_j$, so it can change rankings by penalizing candidates activated
by many bank queries. By contrast, subtracting one mean and dividing by one
standard deviation shared by all candidates for a fixed query cannot change
that query's ranking.

\subsection{Norm-aware scoring}

The inner product is the simplest norm-aware alternative:
\begin{equation}
  s_{\mathrm{dot}}(\bm{x},\bm{y})
  =r_xr_y\cossim(\bm{x},\bm{y}).
\end{equation}
It is appropriate when the product of radii has a validated interpretation, but
it can over-rank vectors with large nuisance norms.

When labeled validation data are available, angle and radius can instead enter
as separate features. For nonzero query and gallery embeddings, a modest
asymmetric score family is
\begin{equation}
  \begin{aligned}
    s_{\bm{\theta}}(\bm{q},\bm{g})
      ={}&\theta_0+\theta_1\cossim(\bm{q},\bm{g})
          +\theta_2\log r_q \\
        &+\theta_3\log r_g
          +\theta_4\left|\log r_q-\log r_g\right|.
  \end{aligned}
  \label{eq:learned_radial}
\end{equation}
This equation is a diagnostic model, not a universally recommended similarity.
For a fixed query, $\theta_0$ and $\theta_2\log r_q$ are constant across gallery
items and therefore cannot change within-query ranking; they matter for
cross-query thresholds or calibration. A probabilistic decision still requires
a fitted link or calibration map. The parameters must be learned on the target
task, regularized, and compared with cosine and inner-product baselines. The
separated coefficients make the claimed role of each radius empirically
falsifiable.

\subsection{Word Rotator's Distance}

Word Rotator's Distance (WRD) is a principled example of using norm and
direction for different purposes at the token level \citep{yokoi2020wrd}.
For a sentence with token embeddings
$\bm{e}_i=\lambda_i\bm{u}_i$, where
$\lambda_i=\norm{\bm{e}_i}$ and $\norm{\bm{u}_i}=1$, define
\begin{equation}
  \nu_s=\sum_i\frac{\lambda_i}{Z_s}\delta_{\bm{u}_i},
  \qquad Z_s=\sum_i\lambda_i.
  \label{eq:wrd_measure}
\end{equation}
Token norms supply transport mass, while angular dissimilarity supplies the
cost
\[
  c(\bm{u}_i,\bm{v}_j)=1-\bm{u}_i^{\mathsf T}\bm{v}_j.
\]
For two sentences with normalized masses $\bm{a}$ and $\bm{b}$, WRD is the
optimal-transport problem
\begin{equation}
  \begin{aligned}
    \operatorname{WRD}(s,t)
      &=\min_{\bm{\Pi}\geq 0}
        \sum_{i,j}\Pi_{ij}c(\bm{u}_i,\bm{v}_j), \\
    \text{subject to}\quad
    \bm{\Pi}\bm{1}&=\bm{a},\qquad
    \bm{\Pi}^{\mathsf T}\bm{1}=\bm{b}.
  \end{aligned}
  \label{eq:wrd}
\end{equation}
Because its ground cost is cosine dissimilarity, WRD is not guaranteed to
satisfy the triangle inequality. The word ``distance'' here follows the
method's name rather than asserting metric status.
WRD is therefore not a radial difference plus one angle between two sentence
vectors. It is a distributional alignment method whose usefulness depends on
token norms being meaningful importance weights.

\subsection{Learned and asymmetric comparison}

When a fixed geometric rule cannot express the task, one can learn the
interaction. A bilinear score
\[
  s_{\bm{M}}(\bm{q},\bm{g})=\bm{q}^{\mathsf T}\bm{M}\bm{g}
\]
can model dimension-specific interactions and, when $\bm{M}$ is not symmetric,
directional relations. A multilayer scorer or cross-encoder is more expressive
still because it allows the two inputs to interact before the final score.
These approaches trade the indexing efficiency and transparency of cosine for
task fit. A common practical compromise is dual-encoder retrieval followed by
a learned reranker.

\section{A Practical Decision Protocol}
\label{sec:protocol}

No scalar diagnostic selects a universally best comparison rule. The following
ordered protocol turns the choice into a reproducible model-selection problem.

\begin{enumerate}
  \item \textbf{Document the representation.} Record model, checkpoint, layer,
  pooling rule, dimensionality, and whether the exposed vector is before or
  after normalization.
  \item \textbf{Identify the training score.} Determine whether the model was
  optimized with cosine, normalized squared distance, inner product, a
  classification loss, or no pairwise objective.
  \item \textbf{Establish equivalent baselines.} Verify vector norms. If they
  are constant, confirm that cosine, dot product, and Euclidean distance give
  identical rankings; do not report them as independent methods.
  \item \textbf{Diagnose the failure.} Measure norm associations, global
  geometry and score margins, neighborhood hubness, and task calibration
  separately. Use a representative reference corpus and avoid test leakage.
  \item \textbf{Apply the smallest targeted remedy.} Test objective-matched
  scoring before post-processing; post-process for global geometry; use local
  or querybank corrections for hubness; and learn radial or asymmetric terms
  only when the task supports them.
  \item \textbf{Evaluate the actual decision.} For retrieval, report recall,
  mean reciprocal rank, or nDCG as appropriate. For semantic textual
  similarity, report correlation with human judgments. For thresholded
  decisions, report discrimination and calibration. Include domain-shift and
  subgroup checks when deployment requires them.
\end{enumerate}

This protocol also clarifies what evidence should accompany a proposed
alternative. A new score should state its invariances, identify the information
it adds or removes, compare against objective-matched baselines, and demonstrate
improvement on the downstream decision rather than on an appealing geometric
visualization alone.

\section{Discussion}
\label{sec:discussion}

\subsection{Similarity belongs to the model--task pair}

An embedding has no unique semantic geometry in isolation. Its coordinates are
the output of an objective, architecture, data distribution, regularization
scheme, and pooling rule. The comparison function adds another inductive bias.
Accordingly, semantic similarity should be treated as a task-defined
relation---paraphrase, topical relatedness, substitutability, relevance,
entailment, or visual correspondence---rather than a universal scalar already
present in every vector space.

This perspective explains both the durability and the limitations of cosine.
It is durable because positive scale invariance is often useful and because
many successful models explicitly train a spherical interface. It is limited
because the same invariance can erase information, while angle alone cannot
repair global distortion, local-density imbalance, task asymmetry, or
miscalibration. The evidence needed for any norm-aware claim is summarized in
Section~\ref{sec:norm}; evidence from one model or layer should not be assigned
to another by analogy alone.

\subsection{Scope and limitations of this review}

This article does not attempt a benchmark-by-benchmark meta-analysis or a
catalogue of every similarity function. It focuses on the recurring conceptual
questions that connect sparse retrieval, representation learning, sentence
embeddings, and cross-modal retrieval. Topics such as hyperbolic embeddings,
kernel learning, probabilistic embeddings, optimal transport beyond WRD,
late-interaction retrieval, and approximate-nearest-neighbor engineering
deserve fuller reviews of their own. The selective scope is intended to make
the decision logic explicit, not to imply that the listed remedies exhaust the
field.

\section{Conclusion}
\label{sec:conclusion}

Cosine similarity earned its central role for good reasons. It offers a clear
geometric interpretation, removes positive radial scale, supports efficient
search, and matches the pairwise interaction used in many normalized training
objectives. Its radial invariance is best justified when the interface fixes
the radius or when norm variation is demonstrably nuisance; the angular
geometry and downstream decision still require validation.

Its limitations are equally precise. Cosine cannot use radial information;
it may be applied to a geometry not identified by the training task; it does
not prevent anisotropy, distance concentration, or hubness; and its symmetric,
uncalibrated score may not match the downstream relation. These failures call
for different remedies. Inner products or explicit norm features address
radial loss. Centering, component removal, whitening, or learned
transformations address global geometry. CSLS and querybank methods address
local-density and hubness effects. Learned scorers and rerankers address
task-specific or asymmetric relations.

The practical conclusion is therefore not that semantics is always angle, nor
that vector norms are always signal. It is that a similarity rule is part of
the model. Its invariances should be stated, its geometry diagnosed, and its
value judged by the decision the system must make.

\balance
\bibliographystyle{unsrtnat}
\bibliography{references_review}

\end{document}